\documentclass[runningheads]{llncs}
\usepackage{listings}
\usepackage{graphicx}
\usepackage{spverbatim}
\usepackage{hyperref}
\begin{document}
\title{Toward accessible comics for blind and low vision readers}
%
%
\author{Christophe Rigaud\inst{1}\inst{2}\orcidID{0000-0003-0291-0078} \and
Jean-Christophe Burie\inst{1}\orcidID{0000-0001-7323-2855} \and
Samuel Petit\inst{2}}
\authorrunning{C. Author et al.}
%
\institute{
L3i Laboratory, SAIL joint Laboratory\\
17042 La Rochelle CEDEX 1, France\\
\email{\{christophe.rigaud,jean-christophe.burie\}@univ-lr.fr}\\
\and
Comix AI (a subsidiary of De Marque Group) \\
5, place des Coureauleurs, 17000, La Rochelle\\
\email{samuel.petit@demarque.com}
}
\maketitle              
\begin{abstract}
This work explores how to fine-tune large language models using prompt engineering techniques with contextual information for generating an accurate text description of the full story, ready to be forwarded to off-the-shelve speech synthesis tools. 
We propose to use existing computer vision and optical character recognition techniques to build a grounded context from the comic strip image content, such as panels, characters, text, reading order and the association of bubbles and characters.
Then we infer character identification and generate comic book script with context-aware panel description including character's appearance, posture, mood, dialogues etc.
We believe that such enriched content description can be easily used to produce audiobook and eBook with various voices for characters, captions and playing sound effects.

\end{abstract}
\section{Introduction}
Visual arts play an important role in cultural life, providing access to social heritage and self-enrichment~\cite{campbell2021reorganizing}. However, most works of art are inaccessible to the visually impaired, whether they are legacy blind, blind, with eye movement disorder or having cognitive eye disease~\cite{Rayar2020ALCOVE}. People with such disease could largely benefit from ``intelligent" computer vision tools in order to also get precise information about visual arts~\cite{sousanis2023accessible} such as comics~\cite{dittmar2014comics}, manga and webtoon~\cite{huh2022cocomix}. Visual details and contexts which are essential to understand and feel the beauty of these artworks are often missing in current experimental tools.


Researchers have considered various ways to make visual art such as comic albums accessible to visually impaired people, including automatic image and/or text/audio descriptions~\cite{ponsard2012ocr} and tactile graphics. 
However, in the systematic review~\cite{oh2021Accessiblity}, Oh \emph{et al.} conclude that image description was out of the scope of interest for most studies, suggesting that automatic retrievals of image-related information is one of the bottlenecks for making images accessible at scale.

To address this challenge, one strategy is to create a comprehensive description of comic books, including layout details, transcriptions of text, and detailed textual descriptions of the graphical content of each panel~\cite{lee2023accesscomics2}. This comprehensive description can then be utilized by text-to-speech tools for playback~\cite{wang2019comic}.
The current state of Text-to-Speech (TTS) models has reached an impressive level of sophistication, allowing them to generate audio that is nearly indistinguishable from human speech.
They are characterized by their ability to produce natural-sounding synthetic voices with emotions, pauses, and realistic tone.
In that scenario, the quality, details and relations of textual transcription and description are key issues for both textual and audio versions of the books. Recent advances using (multimodal) large language models (LLM) based on transformer architecture, sparked a research frenzy by highlighting the impressive capabilities of not only processing text, but also to understand and detail image content~\cite{ramaprasad2023comics}.

In this work, we investigate prompt engineering techniques such as chain-of-prompt to make the most of this technique and dramatically improves prediction quality without requiring additional data or annotation time.
The generated description is intended to mirror the image content by providing an overall scene description with named characters, dialogues and interaction following the natural reading order~\cite{carroll1992visual}.
We propose to use previous computer vision techniques and optical character recognition to extract information from the comic strip images, such as the panels, characters, text, reading order and the association of bubbles and characters. We then use this information as additional context for instruction-tuning large language models and generate a precise description of each panel.
The main contributions of this work are as follows:

\begin{itemize}
    \item Text type classification: sound, caption, dialogue
    \item Automatic character's names inference
    \item Contextual panel description
    \item Script generation including all textual and visual elements    
\end{itemize}

\section{Related works}

In this section, we first review comics accessibility and then comics image analysis.
Finally, we give a focus to recent transformer-based method known as Large Language Model (LLM) applied to comics analysis and understanding.

\subsection{Comics accessibility}

In 2017, Rayar \emph{at al.} called on the document analysis community to address the issues of visually impaired people~\cite{rayar2017accessible}. 
Then several experimental tools have been proposed such as an accessible comic reader for people with low vision~\cite{Rayar2020ALCOVE} and a study on the user experience of accessible comic book reader for textual sound effects~\cite{lee2023accesscomics2,ohnaka2022visual}.
When the participants were asked to select the most important information they desired while reading a comic, the majority prioritized scene descriptions, followed by transcriptions and facial expressions of characters, etc.~\cite{sachdeva2024manga}.
Both studies are compared in~\cite{fontes2022aspectos} (in Spanish).
There is an ongoing collaboration at San Francisco State University about comics studies, a program for visual impairment, and the Accessible Comics Collective which explores ways of making comics accessible for blind and low-vision readers\footnote{https://spinweaveandcut.com/blind-accessible-comics/}.
An approach on Webtoon proposed to utilize comments to improve non-visual Webtoon accessibility. It is based on a panel-anchored adaptive descriptions~\cite{huh2022cocomix}.
Synopsis generation has been explored to have quick and crisp understanding of a comic story~\cite{devi2020cbcs}.
We propose to consider panel-based description, sound effects, script generation and extend them with other text type and character-related information to produce a richer reading experience for Blind and Low-Vision (BLV) readers.


\subsection{Comics image analysis}
\label{sota-comics-analysis}

Recently, a decade systematic literature review has been proposed by ~\cite{sharma2024image} for comics image segmentation, classification and recognition methods.
Earlier, other surveys presented an extended overview of computational approaches for comics analysis~\cite{laubrock2020computational} and for comic research in computer science~\cite{Augereau2018Survey}.
They highlighted numerous methods for most common element extraction like panels, speech balloons and text. Also, some methods have been proposed for very specific elements such as inferring unseen actions~\cite{iyyer2017amazing,vivoli2024multimodal} or balloon tail detection~\cite{lenadora2020extraction,nguyen2019comicMTL,sachdeva2024manga} which are really helpful for accurate text-to-character association~\cite{li2023manga109dialog}.
To our knowledge, automatic text type classification and script generation have not been addressed yet.
We review character clustering and identification (naming) in the two following subsections.

\subsubsection{Character clustering}
\label{character-clustering}

Character clustering (or re-identification) consists in recognizing characters consistently across all different panels in a comics or series of comics.
It is subsequent to character detection for which several methods are proposed in the literature (see surveys introduced in the previous section).
It presents significant challenges due to limited annotated data and complex variations in character appearances~\cite{soykan2023identityaware}.

Initially, k-means based method~\cite{tsubota2018adaptation} was proposed, but they require specifying the number of clusters (character instance groups), which is unknown in our case. To solve this problem, more advanced clustering method which automatically determine the number of clusters (Manga characters) have been proposed~\cite{nguyen2021manga,yanagisawa2020automatic,yanagisawa2019manga,zhang2022unsupervised}.
Ramaprasad \emph{et al.}~\cite{ramaprasad2023comics} used large pre-trained vision encoder Contrastive Language-Image Pre-Training (CLIP)~\cite{radford2021learning} a multi-modal large language model that can perform zero-shot image classification based on natural language prompts.
Sachdeva \emph{et al.} compared their proposition also to CLIP image feature for representing manga characters~\cite{sachdeva2024manga}.
An extended clustering can be achieved by combining visual features with spatial-temporal information, an approach chosen by ~\cite{zhang2022unsupervised} to achieve unsupervised person re-identification in Japanese manga.
Their method relies on design rules such as characters tend to appear on adjacent pages one after another, characters in the same frame tend to belong to different identities, some characters tend to appear in pairs, and so on.
Also, characters’ unique signature is highlighted, such as special accessories, hairstyles, costumes, etc.

\subsubsection{Character identification}


Regarding character identification, Ramaprasad \emph{et al.}~\cite{ramaprasad2023comics} used 
CLIP to predict character's names from a given list of names.
Pre-defined list of name is a strong limitation that we can not afford in our context because character names of little-known or new comics series can not be determined in advance.
From our knowledge, there is only one very recently published method that extend character clustering to character naming and associate them with speech balloon~\cite{li2024zero}.
This method requires characters names list as input and then associate these names to detected character based on nearby text region analysis using LLM.
The ultimate task of matching texts and speakers is challenging in comics analysis, often necessitating an understanding of conversation history and context to disambiguate speakers~\cite{sachdeva2024manga,li2024zero}.

\subsection{Comics and Large Language Model}



Very recently, Sachdeva \emph{et al.} tackled the problem of diarisation i.e. generating a transcription of who said what and when, in a fully automatic way for visual impairment.
The result of this work is similar to our proposition for the dialogue part but does not consider panel scene description.
The work~\cite{li2024zero} proposes to identify and name characters based on a given name list as input and analyse text blocks with LLM.
In our proposition, we extend LLM capacities to automatically build the name list from dialogues.


The work of Ramaprasad \emph{et al.} is the most similar to our contribution by proposing to generate accessible text descriptions for comic strips including panel description using LLM prompt engineering~\cite{sahoo2024systematic} as well~\cite{ramaprasad2023comics}.
In this study, they use famous comics which allows LLM to use their previous knowledge for character name inference but also generate hallucination effects regarding extra details.
MaRU method~\cite{shen2023maru} incorporates a vision-text encoder that combines textual and visual information into a unified embedding space, enabling the retrieval of relevant scenes based on user scene description queries (even in another language).
This method excels in end-to-end dialogue retrieval and exhibits promising results for scene retrieval which enhance the understanding of and improve accessibility of Manga.
To extend this study, the authors propose as future work to recognise comic characters for enhancing attribution and understanding of visual content.
We address the latter in Section~\ref{character-name-inference}.

In~\cite{guo2023m2c}, a method to complement missing comic (manga) text content during the manga creation/translation process is proposed.
This manga argumentation method mines event knowledge within the comics with large language models. Then, fine-grained visual prompts support manga complement.

\section{Proposed method}

The aim of the proposed method is to automatically generate a structured comic book script-like description of each panel including scene, action and dialogues.
This is intended to facilitate enriched text-to-speech (or text-to-braille) conversion for enhanced comic accessibility.
The main challenges are to associate each dialogue to corresponding named characters and generate a detailed text description using natural language.
Our contribution can be seen as one component within a broader pipeline, outlined as follows:



\begin{figure}
\includegraphics[width=\textwidth]{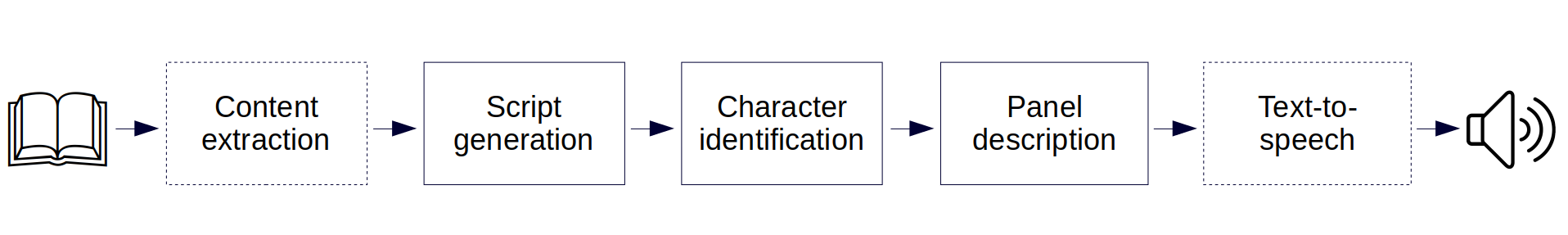}
\caption{Proposed contributions within an example of complete pipeline. Proposed blocks are represented with solid rectangles while complementary ones are dashed.} \label{method}
\end{figure}


\subsection{Script generation}
\label{intermediate-script-generation}

Considering that all the visual and textual elements have been extracted by pre-processing image analysis algorithms, we generate a script of the album gathering all these elements in a single structured text file according to the original layout of each page.
We encode the script using Markdown markup language which is appropriated for both comics script\footnote{https://www.codymarkelz.com/posts/2022-01-20-rmarkdown-comic-script.html} and LLM inference on structured documents~\cite{min2024exploring}.
Then, the script will be  used as context in the next step for further inferences such as character naming (see Section~\ref{character-name-inference}).

\subsubsection{Page layout}

Each comic book page is usually composed by several panels that should be read in a certain order e.g. left-to-right and down the ``Z-path" for English comics books~\cite{cohn2013navigating}.
Assuming that the panel have been previously extracted and ordered during the content extraction pre-processing stage, we propose to list them in a single text file according to their reading order (see script sample at this end of this section).

%

\subsubsection{Text type classification}
\label{text-type-classif}

The script is then completed with textual information from previous content extraction module and inserted into each \texttt{Panel} section of the script following the natural text reading order.
The detection of speech balloon tail with algorithms from the literature allows us to identify most of the spoken text, and we propose a set of rules to also label the onomatopoeia (sound effects) from detected text.
For comics genres that do not make use of tail, balloon contour shape combined with text analysis can be used instead.
We qualify as ``caption" all the remaining text (not classified as spoken or sound effect).
It is important in our study to qualify text as much as possible in order to improve textual description and accessibility (e.g. using different synthetic voices to render different type of text).

\paragraph{Sound effect}
We consider as sound effect (SFX) single text lines with an important height or slope.
If an onomatopoeia is composed of several text lines, they will be considered as separated and different at this stage.
We set the minimum text line height $minH$ - to be considered as sound effect - to $minH = 0.025 \times image  width$ (2.5\%) and the minimum slope $minS$ to $0.1$ (10\%) compared to horizontal line, based on empirical experiments.
Note that these rules may consider a shopfront or other drawing (big or sloped) as onomatopoeia.

\paragraph{Caption}
In comics, caption or narration boxes are used for narration, transitional text (e.g. ``Meanwhile..."), or off-panel dialogue. Captions usually have rectangular borders, but they can also be border-less or floating letters.
Here, we simplify the definition and consider as caption all text blocks that are not classified as sound effect and that haven't been associated to a comic character (no associated tail).
Assuming an error-free result from the previous tail detection algorithm, we expect this category to be composed by usual rectangular text region providing contextual narrative information throughout the story.

\paragraph{Dialogue}
The dialogue sets are composed of all the remaining text, each dialogue being composed by one or more lines of text contained in a speech balloon associated with a comic character (with a tail pointing to it).
Each line of dialogue is preceded by the identifier of the corresponding character i.e. $c0$, $c1$.
Character's identifier computation comes from character clustering which is detailed below.

\subsubsection{Character clustering}

We consider spatial-temporal relationships, as explored in~\cite{zhang2022unsupervised}, 
during the instruction-tuning phase.
This involves using a generated script that contains ordered panels and text as context (see Section~\ref{character-name-inference}).

For character clustering, we compute image feature vectors on each character instance using a variant of Contrastive Language-Image Pre-training (CLIP) associated with Vision Transformer (ViT) like in~\cite{ramaprasad2023comics,shen2023maru}.
This model is designed to perform zero-shot image classification. It has been trained on a large-scale general purpose image dataset and a text dataset.
CLIP uses ViT to get visual features and a causal language model to get the text features. Both the textual and visual features are projected into a latent space of identical dimensions.
In our case, we use only its image embedding part and reduce the dimensionality of its output feature vectors. This simplifies the embedding into a form that traditional clustering methods, as presented here\footnote{https://blog.bruun.dev/semantic-image-clustering-with-clip/}, can handle more effectively.
Then we cluster reduced image feature vectors 
to automatically group the characters by visual similarities, encoded in image vectors.

To reduce the size of image vectors, we use Uniform Manifold Approximation and Projection (UMAP)~\cite{becht2019dimensionality} like in ~\cite{nguyen2021manga} with a size of 5 (see parameter validation in experiment Section~\ref{char_cluster_eval}).
For the clustering, we employed Hierarchical Density-Based Spatial Clustering of Applications with Noise (HDBSCAN)~\cite{mcinnes2017hdbscan} algorithm, which gives state-of-the-art results on comics according to~\cite{nguyen2021manga,yanagisawa2020automatic}.
The minimum cluster size is the only parameter of HDBSCAN, we set it to $15$, assuming that the main characters appear many times in comic books. Groupings smaller than this size will be gathered in a ``noise" group. In our study, we are interested in finding the names of the main characters associated to dialogues. Some less frequent characters might not be named in the story, we expect them to be considers as ``noise" at this stage, and we will refer to them using interrogation mark symbol  ``?" in the script (see Fig.~\ref{fig_char_names}).

Cluster IDs are used to temporarily identify comic characters in the script until they are replaced by inferred proper names in the next step. For example, character $A$ will be identified as $c0$ and character $B$ as $c1$.
We complete the layout-only script with text type information (caption, sound, dialogue) and character clustered IDs as in the following example:
\begin{spverbatim}
    # PAGE 1 - 1 PANEL:
    ## PANEL 1
    ### SOUND
    ### CAPTION
    ### DIALOGUE
    c0: ...
    c1: ...
\end{spverbatim}
We are aware that in our study other type of text might be miscellaneously considered as caption but experiments shown that it doesn't alter character identification and character naming, the next main step objective detailed below.



\subsection{Character's name inference}
\label{character-name-inference}

We propose an automatic method for inferring character names from the generated script.
Contrary to~\cite{ramaprasad2023comics} which provides names manually or other methods that use previous knowledge e.g. famous comic character found on internet, we automatically infer character's name from the script to allows comics accessibility of lesser-known books or new series as well.

To do so, we leverage the inferring capacity of LLM providing our generated script as context~\cite{sahoo2024systematic}.
The script contains spatial-temporal information (panel sequence in reading order and qualified text) and character identifier $c0, c1, etc.$. We propose a chain-of-prompt of four prompts to guide the model throughout its knowledge exploration. LLM is requested to first infer character names using its Named Entity Recognition (NER) capabilities.
In this prompt, we also request the model to explain its reasoning to benefit from chain-of-thought approach and minimize eventual hallucinations~\cite{sahoo2024systematic}.
Secondly, the LLM is used to associate a proper name to each recurring character ID from the clustering by making inferences from dialogue texts. For instance, if there are two characters in a panel and the first character is mentioning a proper name, it is likely the name of the second character, especially if this is repeated in several panels.

Character names are a first step in high-quality accessibility and we propose to go further by characterizing each character with their relationship, genre, age, and temperament in order to, for instance, select or generate the most appropriated synthetic voice for subsequent speech-to-text processing.
We found that many LLM can be used for this task using prompt engineering~\cite{minaee2024large}.
Here is the proposed chain-of-prompt providing an extract of our previously generated script as context in place of \texttt{SCRIPT}:
\begin{spverbatim}
    USER: This is the script of a comic book: """SCRIPT"""
    USER: Please list all character's names.
    USER: Please list all corresponding unique identifiers. Itemize.
    USER: What are their relationship? Explain  your reasoning 
        step-by-step.
\end{spverbatim}



The chain-of-prompt is guiding the LLM from general context such as "a script of a comic book" from which an overall description will be automatically generated according to the general system prompt.
This description forces the network to parse the full script a first time.
Then, the second prompt requests basic information about characters (finding proper names and associated IDs from the script) of the story it just discovered in the previous prompt, avoiding looking for ``farthest" knowledge.
Last, we request their relationship and a last specific information requiring reasoning about character personality.
The assistant output can be easily parsed and character identifier e.g. $c0$, $c1$ are associated with proper names e.g. $Curt$, $Cynthia$ (see Section~\ref{char-inference}).
Relationship details can be used to check how well the model understands the story. It can also be included in a character biography section at the beginning of (accessible) books to introduce characters along with story synopsis.

\subsection{Panel description}
\label{panel-desc}

We propose to include each panel content description in the script to allow the reader to get a better understanding of the story before diving into characters dialogues, such as recommended by the BLV community~\cite{lord2016comics}.

\begin{figure}
\centering
\includegraphics[width=25em]{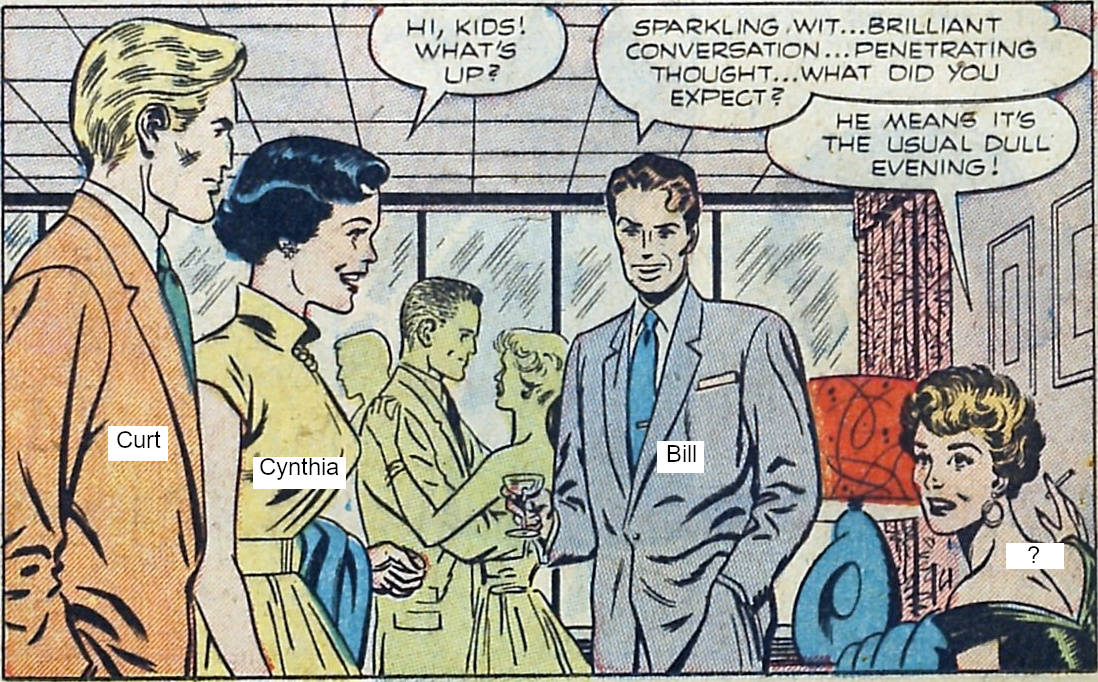}
\caption{Panel 5 of page 12 from Boy Loves Girl 41~\cite{eBDtheque2013Guerin} with identified character's names inserted over corresponding characters.} \label{fig_char_names}
\end{figure}

Image content description has made great strides in the last two years with the rise of Visual Language Models (VLM).
Such models having a great capacity of generalizing, we found that they are able to describe panel images with an important level of detail and give an accurate insight type of lighting, type of view, objects, character, mood, action, etc.
Given that these models can also interpret text~\cite{shen2023maru} like OCR, we introduce character name's identifier (see Section~\ref{character-name-inference}) artificially into panels.
This facilitates correspondences for the VLM between panel images and script content, as illustrated in
Fig.~\ref{fig_char_names}.

Character's names are written as clearly as possible to avoid any text recognition error, using a common font with black colour in a white rectangle.
We position them carefully, ensuring that important details like facial expressions are not hidden.
Our proposal is to centre them on the detected body.

Here is the proposed prompt sequence providing the script as \texttt{SCRIPT}:

\begin{spverbatim}
    USER: Please describe this comic book panel using character's 
        names written in white rectangles.
    USER: Please rephrase considering character's settings and 
        interaction only, naming characters and citing all text
        from the following script extract: """[SCRIPT]"""
\end{spverbatim}
\vspace{1em}
%

We formulate two successive user prompts as shown above.
First, we ask to describe the scene and then rephrase it by including contextual information from the provided script. 
This chain-of-prompt approach~\cite{sahoo2024systematic} helps the network to focus progressively on the elements of interest (mitigate missing information and hallucination effect).
By indicating ``citing all text" in the second prompt, we emphasize the importance of including and referencing all text from the script in the generated description.
The limitation ``settings and character interaction only" mitigate description about art style that models seems very tempted to describe as well e.g. mid-20$^{th}$-century American comics, from the 1950s, etc.

We believe that providing a precise panel description, BLV people will also be able to infer invisible actions that are happening between panels (in the gutters) known as ``closure"~\cite{iyyer2017amazing}.
For example, this is the final description \texttt{DESC1} for the panel presented Fig.~\ref{fig_char_names} given the below \texttt{SCRIPT} extract:\\

Script extract \texttt{SCRIPT} given as context :

\begin{spverbatim}
    ### DIALOGUE
    Cynthia: HI, KIDS! WHAT'S UP?
    Bill: SPARKLING WIT… BRILLIANT CONVERSATION… PENETRATING THOUGHT… 
          WHAT DID YOU EXPECT?
    ?: HE MEANS IT'S THE USUAL DULL EVENING!
\end{spverbatim}
\vspace{1em}

Panel description \texttt{DESC1} that we insert into the script in the corresponding \texttt{PANEL} section:

\begin{spverbatim}
    In this comic panel, Curt, Cynthia, and Bill are having a conversation at a social gathering. Curt, dressed in a brown suit and tie, stands on the left. Cynthia, wearing a yellow dress, stands in the middle. Bill, dressed in a gray suit and tie, stands on the right. The background of the panel features a window with a red curtain and a door. The dialogue between them goes as follows:
    Cynthia: Hi, kids! What’s up?
    Bill: Sparkling wit... brilliant conversation... penetrating thought... what did you expect?
    ?: He means it’s the usual dull evening!
\end{spverbatim}
\vspace{1em}

\section{Experiments}
\label{experiments}





\subsection{Dataset}

For this exploratory research, we limited our experiments to public domain English comics.
A lot of them are available from \href{https://digitalcomicmuseum.com/}{Digital Comic Museum.com/} or \href{https://comicbookplus.com/}{Comic Book Plus} and have already been annotated in several public datasets.
To our knowledge, none of the dataset are proposing text type classification, one dataset (Manga109~\cite{fujimoto2016manga109}) contains character identification annotation and association to text~\cite{li2023manga109dialog} and none of them propose panel description and script.
Note that character clustering and name inference require several (ideally all) consecutive pages from consistent episode, albums or series.
This ensures homogeneous clustering of as many characters as possible at once and the gathering of associated dialogues to facilitate character name inference.
Faced with the lack of appropriated dataset, we chose to manually build and share a toy dataset by considering full albums from which at least some pages were part of eBDtheque dataset~\cite{eBDtheque2013Guerin}.
We extended part of the dataset by getting complementary pages of public domain titles like golden age American comics which are also part of several other public dataset, but not annotated according to our experiments.
We first focused on ``Escape with me" episode of an old title ``Boy Loves Girl 41" (1953), from which eBDtheque pages 12 and 15 are both part of.
This episode is spread oven page 11 to 17 and composed by 45 panels, 75 character instances and 109 
text blocks in total. We exclude advertising content from the last page and do not modify the story analysis.
We downloaded all complementary pages from Comic Book Plus.com\footnote{\href{https://comicbookplus.com/?dlid=31803}{https://comicbookplus.com/?dlid=31803}}.

We also experimented on a second more recent (2003) publicly available comic book: ``Patents" from the World Intellectual Property Organization (WIPO)\footnote{\href{https://www.wipo.int/publications/en/details.jsp?id=67}{https://www.wipo.int/publications/en/details.jsp?id=67}}. It is a free comic book available in English, French, Spanish, Arabic, Chinese and Russian. We chose the English version which is composed of 12 pages, 85 panels, 132 text blocks and 77 character instances.







\subsection{Text type classification}
We evaluated text type classification for each text line generated by an OCR as described in Section~\ref{text-type-classif}.
We used Google Vision API to extract text blocks from images using its Optical Character Recognition (OCR) capacity. We combine it with a balloon contour analysis method~\cite{rigaud2015knowledge} to detect speech balloons and their tail tips.
Each text block detected by the OCR contained by a detected balloon were automatically associated.
Text block not contained by a balloon could then only be considered as sound effect if it has important height or slope as described in Section~\ref{text-type-classif}.
In episode ``Escape with me", all text were detected by the combined text block + balloon contour detection, and we measured the recall (R) and the precision (P) which are 100\% and 93\% respectively.
For ``Patents", $P= 100\%$ and $R=98\%$ respectively.
The precision is not at the maximum because some bubbles were detected as two different text blocks due to their complex shape and illustrative text was detected as balloon (e.g. clock numbers and telephone in the last page of the episode).
We corrected it manually for the rest of the evaluation to avoid error propagation.

Table~\ref{tab1} gives a summary of correct and confused classification for each evaluated episode.

\begin{table}
\centering
\caption{Text type classification confusion matrix comparing reference (ground truth) and predicted text type.}\label{tab1}
\begin{tabular}{|l|l|l|l|l|l|l|}
\hline
& \multicolumn{3}{|l|}{Escape with me} &  \multicolumn{3}{|l|}{Patents} \\
\hline
Ref./Pred. &  Sound & Dialogue & Caption &  Sound & Dialogue & Caption \\
\hline
Sound       & \textbf{1}     & 0     & 0 & \textbf{5} & 1 & 8\\
Dialogue    & 0     & \textbf{77}    & 4 & 1 & \textbf{78} & 0\\
Caption     & 0     & 6     & \textbf{21}& 0 & 18 & \textbf{15}\\
\hline
\end{tabular}
\end{table}

The highest confusion concerns captions that have been classified as dialogue in both experiments.
This is due to a known limit of~\cite{rigaud2015knowledge}: not designed to detect open and thought balloons.
By manually adding a tail to open balloons, we dropped caption false positive down to zero for both titles, which confirms the effectiveness of the proposed method for dialogue and caption text type classification.
Regarding sound effects, the confusion with caption is quite high because we include illustrative text into our defined caption category, despite the fact that these are often depicted as onomatopoeia.

\subsection{Character clustering and name inference}
\label{char_cluster_eval}

\subsubsection{Clustering}
\begin{figure}
    \centering
    \includegraphics[width=20em]{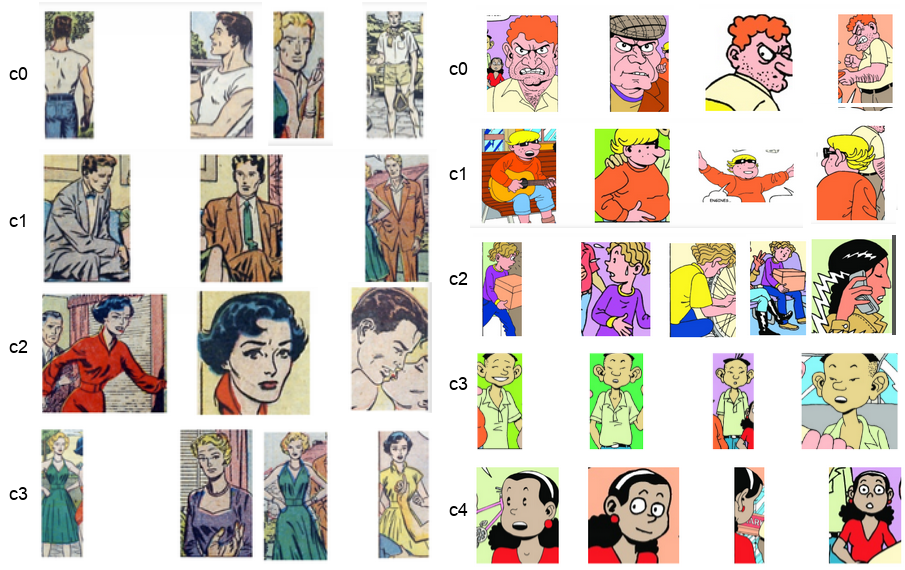}
    \caption{Example of character clustering output: four and five clusters computed for ``Escape with me" (left) and ``Patents" (right) respectively.  
    }
    \label{fig:char-cluster}
\end{figure}

Character clustering is usually preceded by a detection step which supports the consistency in the final textual story reconstruction. To avoid any error propagation, we assume that characters have been detected by any dedicated algorithm from the literature (see Section~\ref{sota-comics-analysis}) and errors are manually curated if needed.
Character crops with significant overlap of multiple character have been discarded to avoid confusion in image feature vector computation and clustering.
For image feature extraction (embedding), we used CLIP ViT-L/14 model trained with the DataComp-1B\footnote{\href{https://huggingface.co/laion/CLIP-ViT-L-14-DataComp.XL-s13B-b90K}{https://huggingface.co/laion/CLIP-ViT-L-14-DataComp.XL-s13B-b90K}} and original hyperparameters similarly to~\cite{guo2023m2c,shen2023maru}.
We varied the minimum cluster size UMAP parameter from 3 to 10 without any notable impact on the final classification. This might be due to the simplistic style of images, we will extend this evaluation to other styles in the future.
We reduced the minimum number of element by cluster of HDBSCAN algorithm to 5 for both titles because of the limited number of pages. Anyway, the algorithm grouped more than 12 instance for each cluster in our experiment, so no impact has been observed.
Fig.~\ref{fig:char-cluster} shows cluster result samples of the groups automatically formed by HDBSCAN plus an extra one (misc) not shown in the figure but detailed in Table~\ref{tab3}.
All clustered images are provided here\footnote{Character set: \href{https://gitlab.univ-lr.fr/crigau02/publication/-/tree/main/2024/MANPU/output/character_set}{https://gitlab.univ-lr.fr/crigau02/publication/2024/MANPU}}.

\begin{table}
\centering
\caption{Character clustering confusion matrix ``Escape with me" and ``Patents".
}\label{tab3}
\begin{tabular}{|l|l|l|l|l|l|l|l|l|l|}
\hline
& \multicolumn{4}{|l|}{Escape} &  \multicolumn{5}{|l|}{Patents} \\
\hline
Ref./Pred. &  c0 & c1 & c2 &  c3 & c0 & c1 & c2 & c3 & c4 \\
\hline
c0      & \textbf{11}     & 3     & 0 & 0 & \textbf{13} & 0& 0 & 0 & 0\\
c1      & 4      & \textbf{10}    & 0 & 0 & 1 & \textbf{10}& 0 & 0 & 0\\
c2      & 1      & 0     & \textbf{14}& 0 & 0 & 0 & \textbf{19}& 0 & 0\\
c3      & 0      & 0     & 6 & \textbf{7} & 0 & 0 & 0 & \textbf{6} & 1\\
c4      & -      & -     & - & - & 0 & 0 & 0 & 0 & \textbf{13}\\
\hline
Misc    & 2      & 4     & 13 & 0 & 2 & 6& 1 & 2 & 3\\
\hline
\end{tabular}
\end{table}

For ``Escape with me" episode, we observed the clustering algorithm as no difficulty to separate male and female characters but mixes sometimes two female characters having only different hair colours or changing clothe a lot e.g. shirtless, in the shadows. 
The misc category contains a lot of $c2$ character, we think that they are mainly due to a lot of back views, from a distance or with face masked for this main character.

The title ``Patents" shows less overall confusion, probably because of better image quality and more colourful characters.
However, the first cluster $c0$ contain in reality two different characters that sometimes wear the same hat and are always together in the story (very close with some body part overlaps and sharing same background).
We observed the same issue for the last character visible in cluster $c2$ (dark-haired woman with straight hair).
She is also very close to the main $c2$ character in the story and her jacket has same colour and zigzag symbols that $c2$ character hairs.
We believe that masking the background could improve the clustering.
Note that these errors have no impact on the character name inference because it is based on dialogue text only.
However, there might be an error propagation during panel description, so we correct them manually before name inference step.

\subsubsection{Name inferences}
\label{char-inference}

Character's name inferences are made from the generated script, as introduced in Section~\ref{character-name-inference}.
In fact, we discovered that LLM have enough reasoning capacity to find protagonist's names from a sequence of ordered and identified dialogues, as long as they are quoted at least once by one of the characters.
Dialogue order and association with comic characters can be computed using automatic methods from the literature (see Section~\ref{sota-comics-analysis}). We used speech balloon and character association method from~\cite{rigaud2015speech} and we measured its accuracy as an indication, assuming that all visible characters have been correctly detected. They are 77.7\% and 88.9\% respectively for “Patents” and ``Escape with me" respectively.
We observed different error types in this method: tail orientation not taken into account (fixed by~\cite{lenadora2020extraction}), balloon with multiple tails, corresponding character not visible (out-of-panel, by phone or in a vehicle, highly overlapping character).
We fixed all wrong associations manually before generating the corresponding scripts used for name inferring, in order to avoid any error propagation effects.

Below is an example of LLM output after applying the chain-of-prompts introduced in Section~\ref{character-name-inference} with the generated script of episode ``Escape with me" containing 80 dialogue type texts.
All tested LLM outputs are available here\footnote{LLM assistant outputs: \href{https://gitlab.univ-lr.fr/crigau02/publication/-/tree/main/2024/MANPU/output/assistant}{https://gitlab.univ-lr.fr/crigau02/publication/2024/MANPU}}:
\begin{spverbatim}
ASSISTANT:
    1. Cynthia Allen (c2) - The protagonist of the story, a young 
    woman from a wealthy family who is struggling to break free 
    from her parents' expectations.
    2. Bill Patterson (c1) - The gardener who Cynthia meets and 
    falls in love with. He is initially portrayed as out of place
    in his job, but later reveals his own dreams and struggles.
    3. Curt (c0) - Cynthia's boyfriend, who represents the life 
    she is expected to lead by her parents.
    4. Cynthia's parents (c4, c3) - They are not named, but are 
    referred to as Cynthia's parents. They want Cynthia to stay
    in Carleton and lead a comfortable, privileged life.
    5. Gloria (c3) - A character who hosts gatherings where 
    Cynthia and Bill first meet. She is portrayed as snobbish 
    and dismissive towards Bill.
    6. The unnamed character (?) - There are a few instances where
    an unnamed character is referenced, likely representing the 
    other guests at Gloria's gatherings.
\end{spverbatim}

Primary character's names are correct in this example: Cynthia (c2), Bill (c1) and Curt (c0).
Secondary character's names ``Gloria" is well predicted as $c3$ but is also wrongly mentioned as not named in the fourth description.
The character $c4$ (Cynthia's dad) is effectively not named in this story. The unnamed characters are representing different character throughout the story (sometimes they are just extras).
We should find a way to differentiate them even if they are not named, and associate them to characters from other panels if they appear many times.

After character's name inference, we replace IDs by names in the script with a simple find-and-replace function, 
The full scripts containing character IDs (before inference) and with IDs replaced by names are provided here for both titles\footnote{Script set: \href{https://gitlab.univ-lr.fr/crigau02/publication/-/tree/main/2024/MANPU/output/script}{https://gitlab.univ-lr.fr/crigau02/publication/2024/MANPU}\label{footnote:script}} named as \texttt{\_script\_1\_} and \texttt{\_script\_2\_} respectively.

\subsection{Contextual panel description}
\label{panel_desc_eval}



Panel image description are generated following the presented chain-of-prompt approach~\cite{sahoo2024systematic} of visual LLM (VLLM) to extract first general visual elements from the scene, then contextualize it with character's names and text from the image and the script.
We randomly selected VLLM from the currently 16 models available in WildVision~\cite{yujie2024wildvisionarena} Arena\footnote{Vision arena: \href{https://huggingface.co/spaces/WildVision/vision-arena}{https://huggingface.co/spaces/WildVision/vision-arena}} to show qualitative results among a large panel of VLLM instead of specific ones.
We processed panel description using at least two different VLLM with slight variation of the user prompt and selected the most accurate descriptions for inclusion in the final script, named \texttt{\_script\_3\_} (see here\footref{footnote:script}).



Note that for this experiment, we manually fixed errors from previous character clustering step (see Section~\ref{char_cluster_eval}) to avoid error propagation in image description.
We evaluated the panel description by measuring the semantic textual similarity with a human annotated panel description for some pages that we use as ground truth.
Text similarity is measured with Sentence Transformers~\cite{thakur2020AugSBERT} implemented in \href{https://www.sbert.net}{sbert.net} library, based on cosine similarity between text embeddings and with the model \emph{mxbai-embed-large-v1}~\cite{emb2024mxbai}.
We evaluated the 14 panel descriptions from the two pages of episode ``Escape with me" available in eBDtheque dataset and also the 12 panels from the two first pages of "Patents".
Concerning the pages from the episode ``Escape with me", the average similarity score is 60.12\% and 69.95\% for the second book.
The evaluation table containing similarity scores for each panel is available here\footnote{\href{https://gitlab.univ-lr.fr/crigau02/publication/-/tree/main/2024/MANPU/output/panel_description}{Evaluation tables: https://gitlab.univ-lr.fr/crigau02/publication/2024/MANPU}}.
We observed that the descriptions depicted character positions, interactions, genres, and attitudes.
The different models considered the information from the provided script extract, and we believe this information plays a major role in the similarity measure.
We should find a more natural way to identify the unnamed speaking characters, especially for subsequent processing such as text-to-speech.
We tried to remove the character label ``?" from the image and some models changed the related part of the description by ``a voice off-panel says: ...". This is not correct in the case of Fig.~\ref{fig_char_names} because the character is visible in the panel, it is just unnamed.



\subsection{Script generation}

The evaluation of the overall quality of the generated script-like description of the story should ideally be done by accessibility experts who are promoting written accessibility\footnote{Accessibility experts: \href{https://www.avh.asso.fr/en}{https://www.avh.asso.fr/en} or \href{https://mangomics-access.fr}{https://mangomics-access.fr}}.
Unfortunately, we have not yet collaborated with such experts to build a public dataset with them. This will be done in a future work.
Another way could be to re-generate an artificial comic book images sequence from the script
and compare it to the original comics version.

\section{Conclusion and future work}

We show that usual comics content analysis combined with zero-shot transformers prompt engineering are paving the way to accessible comics, even without any fine-tuning on comic-specific datasets.
We proposed a straight forward comics script generation based on extracted comics content and enriched with text types, character names and detailed panel description.
The generated script can easily be fed into a text-to-speech module to produce an audiobook and also be used for advanced text-based search and indexing purposes.

In the future, we plan to explore additional text categories for illustrative text, etc. aiming to eliminate confusion with onomatopoeia.
Additionally, we aim to enhance panel descriptions by incorporating content from previous and next panels, thereby improving coherence and avoiding unnecessary repetition.
These enhancements will be validated by accessibility experts.
This research as to be extended to other comics genres and LLM understanding capacities should be challenged across languages using multilingual books for instance.
Page layout and album summary/synopsis information could also be added by extending the chain-of-prompt, according to blind and low-vision people requirements.
For text-to-speech, complementary information such as balloon contour analysis could be used to include speech tone (e.g. shouted, whispered, thought) into the script and modulate the tone and speed of speech synthesis.


%
%
%
\bibliographystyle{splncs04}
\bibliography{mybib}
%




\end{document}